\documentclass{article}
\usepackage{natbib}

\PassOptionsToPackage{compress}{natbib}
\usepackage[dblblindworkshop, final]{neurips_2025}

\usepackage{graphicx} 

\usepackage{amsmath, amssymb}
\usepackage{graphicx}
\usepackage{booktabs}
\usepackage{xcolor}
\usepackage{subcaption}

\usepackage{hyperref}

\hypersetup{
    colorlinks=true,
    linkcolor=blue,
    citecolor=teal, 
    filecolor=green,
    urlcolor=blue}

\title{A Small Math Model: Recasting Strategy Choice Theory in an LLM-Inspired Architecture}
\workshoptitle{MATH-AI}

\author{
  Roussel Rahman\\
  Stanford University\\
  Stanford, CA 94305\\
  \texttt{roussel.rahman@gmail.com}
  \And
  Jeff Shrager\\
  Bennu Climate, Inc.\\
  Palo Alto, CA 94301\\
  \texttt{jshrager@gmail.com}
}

\begin{document}

\maketitle

\begin{abstract}
Strategy Choice Theory (SCT)\footnote{``Strategy Choice Theory'', ``Distributions of Associations'', and ``Overlapping Wave Theory'' have been used to refer to this line of work, emphasizing different aspects.}\citep[e.g.,][]{siegler1984strategychoices, siegler2000rebirth} explains important aspects of children's arithmetic learning based upon principles including learning from developmentally naturalistic data, probabilistic representation, confidence-based retrieval, and the phase-like importance of scaffolding strategies, such as finger-counting. Here we recast SCT as a ``Small Math Model'' (SMM), employing a neural-network-based architecture analogous to LLMs. The SMM extends SCT to include counting practice\footnote{The original SCT model was pre-biased in accordance with the supposed experience of counting.}, symbol (number) embedding, and gated attention. Similar to earlier work, the SMM demonstrates constructive and destructive interference between counting and addition, and the ``wave-like'' use of finger-counting as sum recall improves. We plan to extend the SMM to later aspects of the decades-long SCT program, including adaptive strategy choice and eventually strategy discovery, providing a unified platform to investigate the understanding of numerical characteristics and relationships essential for mathematical reasoning -- as it can emerge in LLM-based agents.
\end{abstract}


\section{Introduction}

In a very real sense, young children participate in reinforcement learning from human (usually parent) feedback. Early theories of children's learning posited processes such as ``stages'' to account for observed differences in children's cognitive skills. This view was overthrown by the work of Robert Siegler and his coworkers \citep[e.g.,][]{siegler1984strategychoices, siegler1995variation, siegler1988strategy, siegler2000rebirth} who proposed the ``Strategy Choice Theory'' (SCT) hypothesizing that cognitive development is a continuous process wherein the child learns to adaptively select from multiple strategies, such as counting, retrieval, finger-counting, and so on, depending upon a set of factors including the strength of association between problem and answer, the child’s confidence, and past experience with each strategy. The SCT framework emphasizes mechanisms that align well with the probabilistic underpinnings, uncertainty estimation, and fallback mechanisms of modern AI. Indeed, LLMs have revolutionized natural language processing by scaling these same principles. 

In the present work, we recast the earliest of the SCT models as a ``Small Math Model'' (SMM) within a contemporary LLM-inspired architecture. In undertaking this work, we have several goals. First, we hope to be able to unify the models that comprise Siegler's early math learning project in an architecture that provides a natural explanation for seemingly diverse phenomena, including learning to count, add, or other arithmetic, learning new strategies, and learning to use strategies in appropriate settings. Second, we hypothesize that the SMM approach can lead to improved ``number common sense'' in LLMs, something that has recently been noted as lacking \citep[e.g.,][]{RahmanMishra2025FragileNumberSense}. 
With similar goals, \citet{hein2024exploring} examined how multimodal transformers acquire skills on a set of numerical tasks (such as enumeration, set comparison, and seriation). They are, however, focused on learning from visual and linguistic data and integrating these modalities, but not on learning the numerical symbols, operations, and their internal representations of magnitudes, which form a crucial part of human number sense, alongside visual and linguistic representations.\footnote{The triple code model \citep{dehaene1992varieties} explains that humans represent numbers in three forms: (1) visual (e.g., $31$), (2) verbal ("thirty-one"), and (3) their analog magnitudes along a number line, each corresponding to different parts of the brain \citep{dehaene1996organization, kiefer1997time, pinel1999event}.} 

SCT has long been used to explain how humans develop number sense \citep{dehaene1992varieties}, making the SMM an ideal testbed for investigating and improving number sense in AI agents. While modern XAI methods, such as mechanistic interpretability, have yielded insights into how neural networks solve mathematical problems by rebuilding solution strategies from elements \citep[e.g., ][]{wang2022interpretability, nanda2023progress}, existing approaches are largely observational and difficult to scale \citep{bereska2024mechanistic}. In contrast, the SMM offers a controllable environment for studying complex reasoning, with clear ground truths about the problem environment and model capabilities.

\section{Strategy Choice Theory}

In complex environments, learners have access to a variety of problem-solving strategies. SCT explains how individuals select among these strategies based on the distribution of associations between problems and potential answers. For example, children learning to add or multiply do not memorize facts (e.g., $2 + 3 = 5$, $2 \times 3 = 6$) but, like an LLM, build a probabilistic ``Distribution of Associations'' between problems and possible answers. Early on, these associations are noisy, as children might associate $2 + 3$ with answers $3$ (counting up from $2$), $4$ (counting from $3$, or $2\rightarrow3$), $5$ (the correct answer), or even other results that they may have encountered through errors early on in their learning to add. With experience, correct associations strengthen while incorrect ones weaken. The SCT is formalized with three mechanisms:  (1) representation, (2) solution generation (through retrieval or fallback strategies), and (3) learning. When retrieval confidence falls below a threshold, the agent resorts to backup strategies and uses it as a learning opportunity to update associations. 

\section{Architecture}
\label{sec:architecture}

The SMM processes input sentences of the form ``$a\ operator\ b$'' where $a,b \in \{1,\ldots,5\}$ and $operator \in \{\rightarrow (count\ up), +\ (add)\}$. The symbolic numerical arguments (i.e., the numbers) are ``embedded'' or mapped to a dense vector in a continuous space, enabling the model to capture emergent relationships among numbers (for example, relative magnitude, even versus odd, or common configurations such as base-five hand groupings). These embeddings feed into a recurrent processing core that implements a mixture of strategies, including finger counting and direct addition.\footnote{While the network itself is feed-forward, the training regime and task structure make it functionally recurrent as its own outputs are fed back as inputs across steps to implement counting and addition.} A gating (‘quasi-attention’, QA) mechanism learns how much attention to pay to the operands, conditioned on the operator, so that it emphasizes different input features in different contexts. Faced with a counting problem, it learns to amplify the representation of the current number in the sequence ($b$), facilitating stepwise incrementing, whereas faced with an addition problem, it learns to highlight both operands simultaneously. The embeddings interact with the attentional gate, enabling the model to capture the relational structure of numbers vis-\`a-vis strategies, and vice versa.

Like the 1984 SCT, the SMM includes two strategies for addition: direct recall and finger-counting. It also includes a mechanism for learning to count up, through either recall or ``oracular'' knowledge (being told, presumably by a parent), which was not modeled in the 1984 work. Answer confidence for either counting or addition is estimated from the entropy of the output distribution. When confidence falls below a threshold, the model defers to an alternative strategy for addition, either being told (for counting) or finger-counting, which generates the correct result step by step. Importantly, the SMM uses its counting skills as a scaffold for learning addition skills. The finger-counting strategy relies on the ability to count up, which is learned by the model itself! The subsequent answer is then used for training in the usual back-propagation way. This is analogous to the way LLMs can be augmented with external tools (e.g., for coding) or explicit reasoning mechanisms (e.g., chains-of-thought, \citeauthor{wei2022chain}, \citeyear{wei2022chain} or tree-of-thoughts, \citeauthor{yao2023tree}, \citeyear{yao2023tree}) and subsequently distilled back into the base model.

\section{Training}
\label{sec:training}

Training follows a Gaussian curriculum schedule that gradually increases task difficulty. Early in training, simpler problems (e.g., sums with smaller operands) dominate; over time, harder problems appear. This smooth progression mirrors natural learning trajectories while avoiding abrupt jumps. It also follows learning practices in LLM training, where data distributions are staged over time to improve stability and generalization.

Counting is trained from the very beginning of each run, while addition can be introduced at different time points to investigate how prior knowledge influences subsequent learning. After each problem, strategy-specific confidence values are updated according to correctness, and these values govern stochastic selection of strategies on subsequent trials. This architecture, therefore, supports mechanistic analysis of how new procedures compete with, reinforce, or interfere with existing ones in the course of skill acquisition. The code for the SMM is available here (under MIT Open Source License): \url{https://github.com/jeffshrager/smm} 

\section{Results}

\begin{figure}[b]
  \centering
  \begin{subfigure}{0.48\linewidth}
    \centering
    \includegraphics[width=\linewidth]{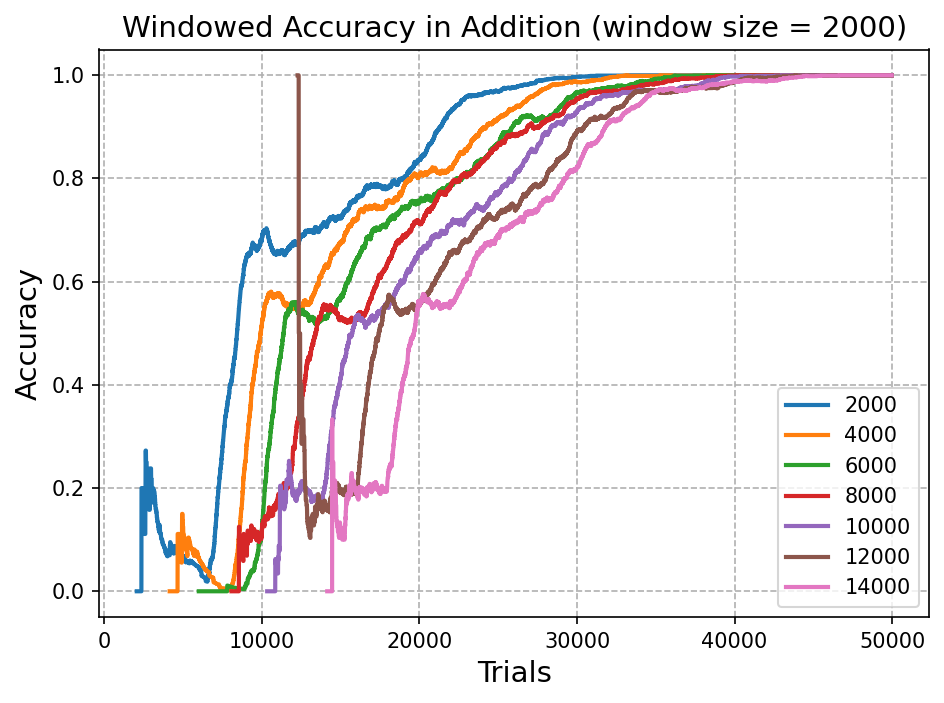}
    \label{fig:1a}
  \end{subfigure}
  \hfill
  \begin{subfigure}{0.48\linewidth}
    \centering
    \includegraphics[width=\linewidth]{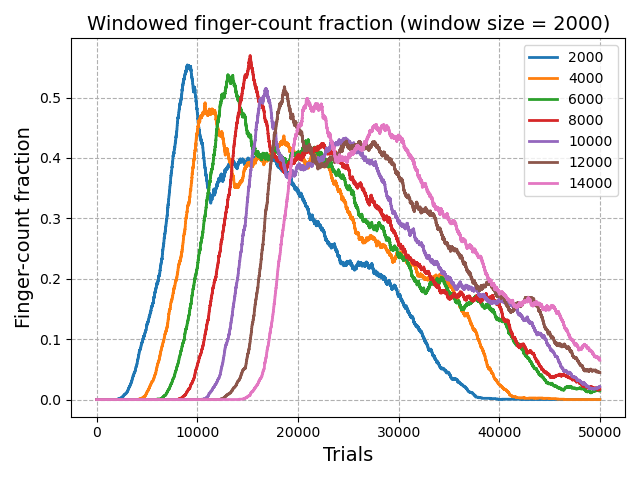}
    \label{fig:1b}
  \end{subfigure}
  \caption{Addition correctness (a) and use of finger-counting (b) as a function of training step, for different
  addition-start times. Counting is always trained from the beginning. Destructive interference takes place near the beginning of practice if addition is introduced too early, although this is rapidly overcome by training.}
  \label{fig:1}
\end{figure}

Prior experience with counting can either scaffold or hinder the acquisition of addition, depending upon the timing of addition's introduction. Figure~\ref{fig:1} shows changes in addition accuracy with training (Fig. \ref{fig:1}a) and in the use of the finger-counting strategy (Fig. \ref{fig:1}b) for addition problems. We varied the points in training at which addition problems were introduced. In all cases, we see that the change-points in accuracy closely align with the ones in finger-counting usage. The accuracy curves begin with plateaus (of varying lengths) before a rapid rise; during this period, finger-counting usage rises to a maximum, indicating that the model becomes more confident that finger-counting is the more promising strategy and that improved addition skills emerge from improved counting skills. Then follows another brief plateau, followed by a gradual rise to near-perfect accuracy. During this period, finger-counting usage briefly reached several local maxima, reflecting the continued use of finger-counting as a learning strategy, before a gradual decline to zero, marking the end of finger-counting as a scaffold for addition. 
 
\begin{figure}[t]
\centering
\includegraphics[width=\textwidth]{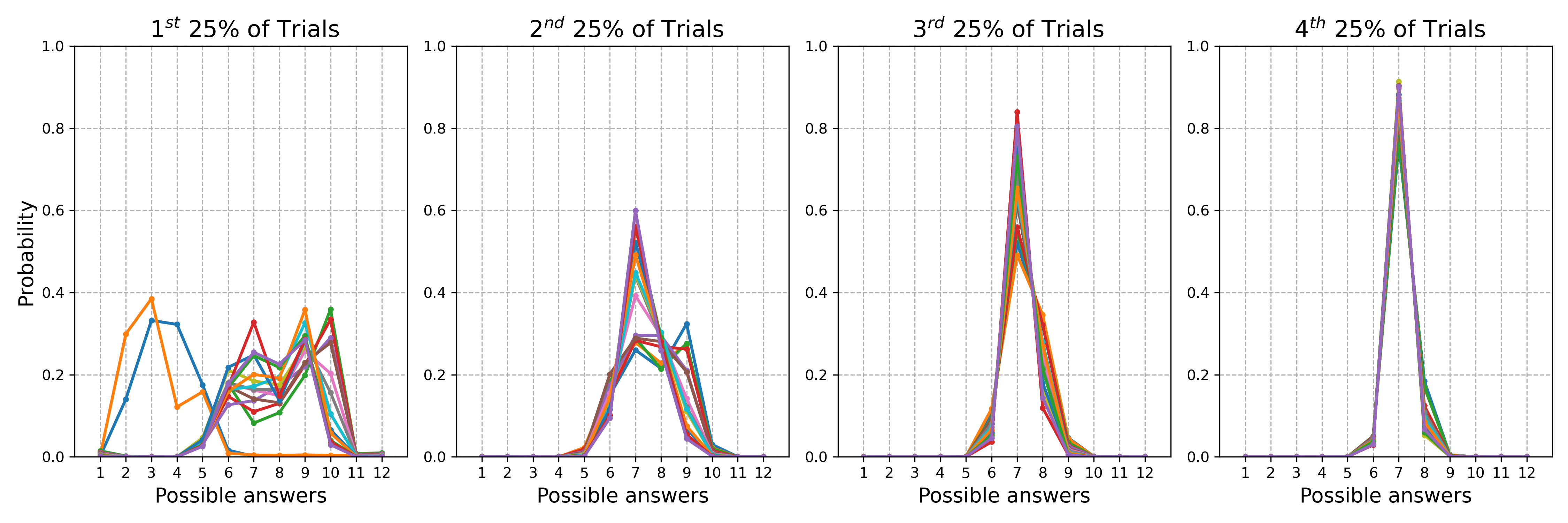}
\caption{Addition answer confidence over the whole 50,000 training cycles, for the problem $3+4=?$, depicted in four quarters, early through late. In each quarter, the lines represent distributions for problems sampled once in every 10 instances. (The training itself is not divided into four periods.)} \label{fig:threeplusfour} 
\end{figure}

The change with learning in addition answer confidence is depicted in figure~\ref{fig:threeplusfour} for a specific problem: $3 + 4 = ?$. Note that early on the possible answers are biased to incorrect results due to addend-bias (for the problem $3+4$ tend to say $3$ or $4$), counting bias (tend to say $5$, from the counting sequence $3\rightarrow 4 \rightarrow 5$), and errors in executing finger-counting due to poor counting training (i.e., this many answers around, but not at $7$, rather at $6,8,9,...$). Notice as well that confidence is distributed among these various answers. Yet, with experience, these biases fall off, and the correct answer ($7$) becomes nearly the only likely answer with nearly 1.0 confidence. 

\section{Discussion and Next Steps}

The SMM modernizes and extends the SCT \citep[][]{siegler1984strategychoices, siegler1988strategy} recasting it in a contemporary LLM architecture that includes embeddings, gating/attention, cross-entropy loss, curriculum learning, and tool (strategy) use. The simple SMM architecture provides a testbed for studying how counting priors, confidence estimation, and quasi-attention (in the LLM sense) interact in a realistic yet controlled setting. We show that the SMM can build upon its counting skills to learn addition, using counting as a scaffold during early training. The transition point—when counting becomes redundant for addition—is clearly identifiable and aligns with predictions from Strategy Choice Theory (SCT), emerging naturally from the model’s confidence thresholds. Moreover, the evolution of confidence distributions over possible answers is interpretable and reveals how biases (e.g., addend or counting biases) diminish with experience. These dynamics not only validate SCT’s core mechanisms but also demonstrate how modern neural architectures can yield transparent, mechanistic insights into the development of symbolic reasoning. 

While our demonstrations focused on problems involving numbers 1 through 5, the SMM architecture is readily scalable to larger numerical ranges and additional operations (e.g., subtraction, multiplication, division), with computational cost being the primary constraint---two directions we aim to pursue to investigate how complex mathematical skills emerge from foundational numerical abilities. Importantly, adaptive strategy competition can give rise to nontrivial transfer effects. In the early part of learning addition, the SMM falls back on its counting skills for solutions and updates associations between problems and answers. When addition problems were introduced too early, the SMM had difficulty using finger-counting to scaffold weak addition skills. An important implication of these results is the potential value of ``progressive deepening'' \citep[e.g.,][]{fayek2020progressive, hacohen2019curriculum}, also recognized in cognitive science as the ``starting small'' hypothesis \citep{elman1996rethinking}. By analogy, staged curricula in LLMs may help establish core abilities---such as number sense---before introducing higher-level reasoning, supporting more robust common sense in numerical domains.

The SMM may further serve as a testbed to probe number sense, which is fundamental to mathematical reasoning. The embedding-based architecture has the potential to capture latent properties (e.g., even/odd or base-five hand configurations, and many others). Indeed, our immediate next steps will be to provide a number of strategies that vary in their effectiveness -- combining efficiency and correctness -- for different sorts of addends.\footnote{Efficiency and correctness are partly correlated because strategies that require more steps have a greater likelihood of error, although there are other factors at play as well.} We expect number embeddings to be able to learn the relationships between number properties and strategy effectiveness. 

The SMM illustrates how modern neural implementations can instantiate and extend classical cognitive theories. By reframing \citeauthor{siegler1984strategychoices}'s model (and eventually the rest of the Siegler line of research) in a modern architecture, we hope to obtain a platform for mechanistic analysis of symbolic reasoning in tractable domains, opportunities to probe representations directly (e.g., via similarity analysis or attention visualization), and a rich bridge between cognitive science and AI, highlighting shared principles of probabilistic selection based on confidence estimation, and adaptive strategy use.

\section*{Acknowledgments}

The authors thank David William Braithwaite for useful comments on a draft of this paper and Michael J. Schoelles for sharing his immense knowledge about human mathematical reasoning.

\bibliographystyle{apalike}
\bibliography{math_reasoning_bibliography}

\end{document}